\documentclass[10pt,twocolumn,letterpaper]{article}

\usepackage{cvpr}
\usepackage{times}
\usepackage{epsfig}
\usepackage{graphicx}
\usepackage{amsmath}
\usepackage{amssymb}
\usepackage{booktabs}
\usepackage{multirow}
\usepackage{bm}

\usepackage[breaklinks=true,bookmarks=false]{hyperref}

\cvprfinalcopy 

\def\cvprPaperID{1335} 

\ifcvprfinal\fi
\begin{document}

\title{Sparsifying Neural Network Connections for Face Recognition}

\author{Yi Sun$^{1}$~~~~~~~~~~~~~~~~~~~~~Xiaogang Wang$^{2,4}$~~~~~~~~~~~~~~~~~~~~~Xiaoou Tang$^{3,4}$\\
{\small $^1$SenseTime Group}\\
{\small $^2$Department of Electronic Engineering, The Chinese University of Hong Kong}\\
{\small $^3$Department of Information Engineering, The Chinese University of Hong Kong}\\
{\small $^4$Shenzhen Institutes of Advanced Technology, Chinese Academy of Sciences}\\
{\tt\small sunyi@sensetime.com~~~~~xgwang@ee.cuhk.edu.hk~~~~~xtang@ie.cuhk.edu.hk}
}

\maketitle

\begin{abstract}
   This paper proposes to learn high-performance deep ConvNets with sparse neural connections, referred to as sparse ConvNets, for face recognition. The sparse ConvNets are learned in an iterative way, each time one additional layer is sparsified and the entire model is re-trained given the initial weights learned in previous iterations. One important finding is that directly training the sparse ConvNet from scratch failed to find good solutions for face recognition, while using a previously learned denser model to properly initialize a sparser model is critical to continue learning effective features for face recognition. 
   This paper also proposes a new neural correlation-based weight selection criterion and empirically verifies its effectiveness in selecting informative connections from previously learned models in each iteration. When taking a moderately sparse structure ($26\%$-$76\%$ of weights in the dense model), the proposed sparse ConvNet model significantly improves the face recognition performance of the previous state-of-the-art DeepID2+ models given the same training data, while it keeps the performance of the baseline model with only $12\%$ of the original parameters.
\end{abstract}

\section{Introduction}

The number of parameters in a deep neural network is restricted by the amount of training data. To reduce model parameters, we introduce a new weight sparsifying algorithm for deep convolutional neural networks, and the learned models are referred to as sparse ConvNets. The sparse ConvNet is derived from a baseline high-performance VGG-like deep neural network \cite{simonyan2014}. When trained on the same approximately $300,000$ face images as DeepID2+ \cite{sun2015}, the baseline VGG-like model achieves $98.95\%$ face verification accuracy on LFW \cite{huang2007a} taking an entire face region (and its horizontally flipped counterpart) as input. When the sparsity is introduced to this baseline model, we could significantly improve the performance from $98.95\%$ to $99.30\%$, reducing the error rate by $33\%$. Moreover, there is a trade-off between model sizes and the performance, and the performance of our baseline model can be kept with only $12\%$ of the original model sizes/parameters. A small model size is preferred on some platforms such as mobile devices.

\begin{figure}[t]
\begin{center}
\includegraphics[width = \linewidth]{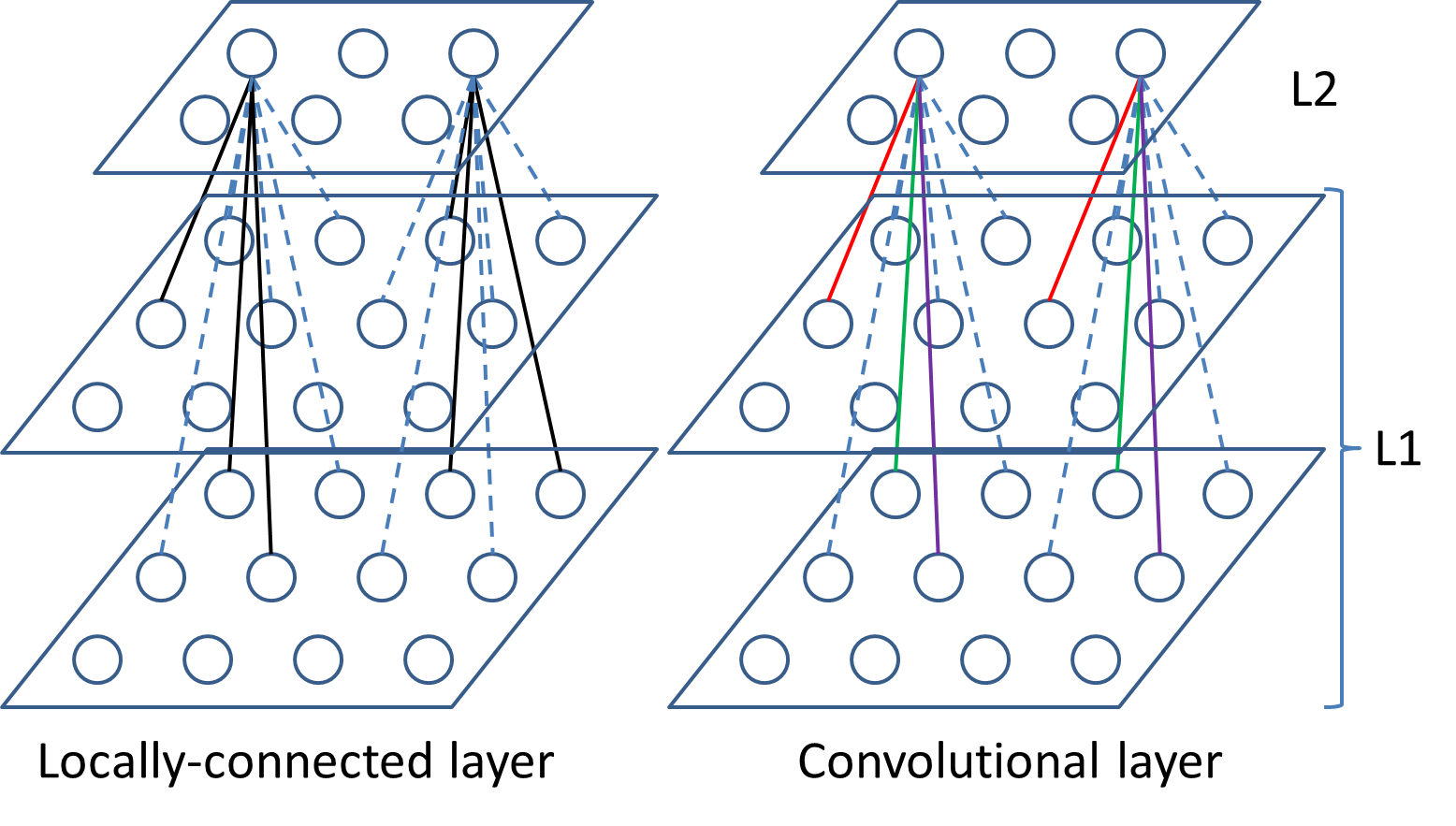}
\end{center}
\vspace{-12pt}
\caption{Weight/connection pruning in a locally-connected layer (left) and a convolutional layer (right). Connections of two neurons in layer L2 are shown while those of the other neurons are omitted. The solid and dashed lines indicate the reserved and pruned connections, respectively. Lines in the same color in the convolutional layer indicate connections with shared weights. Best viewed in color.}
\vspace{-2pt}
\label{fig:demo}
\end{figure}

The idea of reducing neural connections has been taken in designing GoogLeNet \cite{szegedy2015}, which achieved great success on the ImageNet challenge \cite{russakovsky2015}. GoogLeNet reduced neural connections by using very small convolution kernels of sizes $1\times1$ and $3\times3$. We further improve the degree of sparsity by dropping connections in the already very small $3\times3$ convolution kernels in our base model and dropping across different input feature maps. Moreover, the degree of sparsity in our sparse ConvNets can be well controlled by a single sparsity parameter, which makes it easier to make the tradeoff between the performance and model sizes.

Fig. \ref{fig:demo} illustrates weight pruning in a locally-connected layer (layer L2 in the left) and a convolutional layer (layer L2 in the right), respectively. Each neuron is layer L2 is connected to a $2\text{ (width)}\times2\text{ (height)}\times2\text{ (channels)}$ local region in the previous layer (layer L1). Pruning is performed on connections connected to the same input channel as well as on connections across different channels. The set of connections of each neuron in the convolutional layer are pruned consistently due to weight sharing.

Inspired by the Hebbian rule that ``neurons that fire together wire together'' \cite{arora2013}, we drop more connections between weekly correlated neurons than those between strongly correlated neurons. The correlation between two connected neurons are defined by the magnitude of the correlation between their neural activations. On the other hand, neurons in the previous layer which are more correlated (either positively or negatively) to a given neuron in the current layer are more helpful to predict the activities of the latter.

We first train the baseline convolutional model, and then dropping connections layer-wisely from the last to the previous layers, each time only one additional layer is sparsified and the entire model is re-trained. The previously trained models are used to calculate the neural correlations and initialize the subsequent sparser models. It is found that directly train a sparsely connected model is difficult and lead to inferior performance. We conjecture this is because without the help of denser models, a sparse model could easily get stuck to bad local minimums. A denser model with more parameters have more degrees of freedom to avoid such local minimums in the initial training stages. Therefore denser models could provide better initialization for sparser models.

\section{Related work}

Removing unimportant parameters in deep neural networks was studied by LeCun \etal \cite{lecun1990} in their seminal work Optimal Brain Damage. They took a second derivative-related criterion for removing parameters. The second derivatives of parameters are calculated efficiently (but approximately) by back-propagation. They reduced model parameters by four to eight times without loss of the prediction ability of the original model. Optimal Brain Surgeon \cite{hassibi1993} and \cite{srinivas2015} took an additional surgery step when a parameter is pruned to adjust the remaining parameters. In \cite{williams1994,collins2014}, neural weights are regularized by $l_p$ norm (\eg, $p=0,1,2$) and weights with small magnitudes are pruned. Blundell \etal \cite{blundell2015} treated neural weights as Gaussian random variables and estimated the means and variances of weights. Weights with small means and large variances are pruned. Neural networks in \cite{hassibi1993,srinivas2015,blundell2015} did not need to be fine-tuned after pruning. In contrast to the previous studies, we investigate a new weight pruning criterion which explores correlations between neural activations. Through the study on the challenging face recognition problem, it is shown that neural correlations are better indicators of the significance of neural connections than weight magnitudes or second derivatives of weights. Another important finding is that learning an initial dense model is critical to the following learning of sparser models.

Orthogonal to weight pruning, \cite{denton2014,jaderberg2014,zhang2015} explored singular value decomposition and low rank approximation of neural layers for model compression. \cite{ba2014,hinton2014} proposed knowledge distillation, in which a small model (a single model) is trained to mimic the activations of a large model (an ensemble of models). Our weight pruning method may be combined with these techniques. For example, a small model may be first learned with knowledge distillation. Then weights in the small model is further pruned according to some significance criterion.

\section{Baseline model}
\label{sec:base}

Our baseline model is similar to VGG net \cite{simonyan2014} with every two convolutional layers following one max-pooling layer. One major difference is that the last two convolutional layers are replaced by two locally-connected layers. The aim is to learn different features in different face regions, since face is a structured object, and local connections increase the model fitting ability. The second locally-connected layer is followed by a $512$-dimensional fully-connected layer. The feature representation in the fully-connected layer is used for the following face recognition. Tab. \ref{tab:base} shows the detailed structure of our baseline model.

Joint identification-verification supervisory signal \cite{sun2014c} is added to the last fully-connected layer to learn a feature representation discriminative to different face identities while consistent for face images of the same person. The same supervisory signal is also added to a few previous layers to enhance the supervision in previous feature learning stages \cite{sun2015}. Rectified linear activation function \cite{nair2010} is used for all convolutional, locally-connected, and fully-connected layers. Dropout learning \cite{hinton2012} with $30\%$ and $50\%$ dropout rates are used for the last locally-connected and fully-connected layers, respectively, during training.

When the training set is moderately large, our baseline model has achieved the highest face verification accuracy on LFW compared the state-of-the-art methods.
For example, when trained on the same approximately $300,000$ training face images as DeepID2+ \cite{sun2015}, our single baseline model taking the entire face region as input achieves $98.95\%$ face verification accuracy on LFW \cite{huang2007a}, compared to $98.70\%$ for a single DeepID2+ model \cite{sun2015}. The improvement over DeepID2+ is mainly due to larger input dimensions ($112\times96$ compared to $55\times47$) and increased model depth. While the recently proposed FaceNet \cite{schroff2015} has obtained the highest verification accuracy ($99.63\%$) on LFW, it required around $200$ million training samples (almost $700$ times large than ours).
The extremely large train data required makes it impossible to be reproduced by us, and we did not choose FaceNet as the baseline.

\begin{table}[t]
\footnotesize
\begin{center}
\begin{tabular}{p{2.5cm}|p{1.3cm}|p{2.0cm}|p{1.1cm}}
\toprule
type & patch size/ stride & output size & params \\
\midrule
convolution (1a) & $3\times3/1$ & $112\times96\times64$ & 1.8K \\
convolution (1b) & $3\times3/1$ & $112\times96\times64$ & 37K \\
max pool & $2\times2/2$ & $56\times48\times64$ & \\
convolution (2a) & $3\times3/1$ & $56\times48\times96$ & 55K \\
convolution (2b) & $3\times3/1$ & $56\times48\times96$ & 83K \\
max pool & $2\times2/2$ & $28\times24\times96$ & \\
convolution (3a) & $3\times3/1$ & $28\times24\times192$ & 166K \\
convolution (3b) & $3\times3/1$ & $28\times24\times192$ & 332K \\
max pool & $2\times2/2$ & $14\times12\times192$ & \\
convolution (4a) & $3\times3/1$ & $14\times12\times256$ & 443K \\
convolution (4b) & $3\times3/1$ & $14\times12\times256$ & 590K \\
max pool & $2\times2/2$ & $7\times6\times256$ & \\
local connection (5a) & $3\times3/1$ & $5\times4\times256$ & 11.8M \\
local connection (5b) & $3\times3/1$ & $3\times2\times256$ & 3.5M \\
full connection (f) & & $512$ & 786K \\
\bottomrule
\end{tabular}
\end{center}
\vspace{-4pt}
\caption{Baseline ConvNet structures.}
\vspace{0pt}
\label{tab:base}
\end{table}

\section{Sparse ConvNets}
\label{sec:scnn}

Our starting point is the high-performance well trained baseline model $N_0$ as described in Sec. \ref{sec:base}. Then we delete connections in the baseline model in a layer-wise fashion, from the last fully-connected layer to the previous locally-connected and convolutional layers. When a layer $L_m$ is sparsified, a new model $N_m$ is re-trained initialized by its previous model $N_{m-1}$. Therefore, a sequence of models $\{N_1, \ldots, N_M\}$ with fewer and fewer connections are trained and $N_M$ is the final sparse ConvNet obtained.
During the whole training process, the previously learned model is used to calculate the neural correlations and guide the connection dropping procedure. The weights learned by the denser model $N_{m-1}$ are also good initialization of the sparser model $N_{m}$ to be further trained. We first delete connections in higher layers because the fully- and locally-connected layers have the majority of parameters in the deep model. It is found that the large amount of parameters in these layers have a lot of redundancy. Parameters in these layers could be greatly reduced without degrading the performance.

The sparser model could be easily trained by existing deep learning tools such as Caffe \cite{jia2014}. We use a binary matrix (referred to as dropping matrix) of $0$s and $1$s with the same size as the weight matrix of a layer to specify the dropped or reserved weights of the given layer. Each time before forward-propagation, weights in the given layer are first updated by dot-multiplying the dropping matrix. Then the following forward- and back-propagation operations could be done in the same way as a normal denser model, while the model would behave as a sparsely-connected one. The dropped weights being updated after back-propagation would be clipped to zero again before next forward-propagation. At the test stage, the code could be particularly optimized for the sparsely connected layers to actually speedup the computation and reduce the model size for storage, when the model is implemented in various platforms and devices. Even with the current implementation, the model size has already been largely reduced, since the storage of binary dropping matrices is much smaller than that of dropped real-valued weights.

The training algorithm is summarized in Tab. \ref{tab:algorithm}, in which the weight sparsifying criteria will be described in Sec. \ref{ssec:drop}.

\begin{table}[t]
\small
\begin{center}
\begin{tabular}{p{220pt}}
\toprule
\textbf{input}: network structure $T$; layers to be sparsified $L_1$, $L_2$, ... ,$L_M$; degrees of sparsity $S_{L_1}$, $S_{L_2}$, ... ,$S_{L_M}$ \\
\midrule
train baseline network $N_0$ with structure $T$ \\
\textbf{for} $m$ from $1$ to $M$ \textbf{do} \\
\quad calculate dropping matrix $D_{L_m}$ of layer $L_m$ according to \\
\quad the neural correlations in network $N_{m-1}$ and the sparsity \\
\quad degree $S_{L_m}$ \\
\quad initialize network $N_m$ with structure $T$ and weights of \\
\quad network $N_{m-1}$ \\
\quad \textbf{while} not converge \textbf{do} \\
\quad \quad update weights in layers $L_1$, $L_2$, ... ,$L_m$ by dot- \\
\quad \quad multiplying them with dropping matrices $D_{L_1}$, $D_{L_2}$, ... \\
\quad \quad , $D_{L_m}$, respectively \\
\quad \quad forward- and back-propagation one mini-batch of train- \\
\quad \quad ing samples in network $N_m$ and update weights in \\
\quad \quad network $N_m$ \\
\quad \textbf{end while} \\
\textbf{end for} \\
\textbf{output} network $N_M$ with sparsified connections specified by \\
dropping matrices $D_{L_1}$, $D_{L_2}$, ... ,$D_{L_M}$ \\
\bottomrule
\end{tabular}
\end{center}
\vspace{-4pt}
\caption{The sparse ConvNets learning algorithm.}
\vspace{0pt}
\label{tab:algorithm}
\end{table}

\subsection{Sparsify connections}
\label{ssec:drop}

Given the degree of sparsity $S$ ($0<S<1$), we sample $S\cdot |W|$ weights from the total number of weights $|W|$. The number of connections are proportional to the number of weights for all types of layers in our deep ConvNets. The sampling is based on neural correlations. In principle, we tend to keep connections (and the corresponding weights) where neurons connected have high correlations and drop connections between weakly correlated neurons. This is because neurons in one layer which have stronger correlations to neurons in the upper layer have stronger predictive power for the activities of the latter. Note that neurons with strong negative correlations are also useful in predicting neural activations. If a neuron is viewed as a detector of a certain visual pattern, its positively correlated neurons in the lower layer provide evidence on the visual pattern, while its negatively correlated neurons help to reduce false alarms.
In practice we find that keeping a small portion of connections to weakly correlated neurons is also helpful. We conjecture the reason might be that predictions from weakly correlated neurons are complementary to those from highly correlated neurons.

First consider fully- and locally-connected layers in which weights are not shared. Weights and connections are one-to-one mapped in these layers. Given a neuron $a_i$ in the current layer and its $K$ connected neurons $b_{i1}$, $b_{i2}$, ... ,$b_{iK}$ in the previous layer, the correlation coefficient between $a_i$ to each of $b_{ik}$ for $k=1,2,\ldots,K$ is (for simplicity, when we refer to a neuron, we also mean its neural activations)

\begin{equation}
r_{ik} = \frac{E[a_i - \mu_{a_i}][b_{ik} - \mu_{b_{ik}}]}{\sigma_{a_i}\sigma_{b_{ik}}} \ \textrm{,}
\end{equation}

where $\mu_{a_i}$, $\mu_{b_{ik}}$, $\sigma_{a_i}$, and $\sigma_{b_{ik}}$ denote the mean and standard deviation of $a_i$ and $b_{ik}$, respectively, which are evaluated on a separate training set. Since both positively and negatively correlated neurons are helpful for the predictions, we consider the corresponding connections respectively. From all $r_{ik}$ for $k=1,2,\ldots,K$, we first take out all positive correlation coefficients and sort them in descending order, denoted as $r_{ik}^+$ for $k=1,2,\ldots,K^+$. Then we randomly sample $\lambda SK^+$ and $(1-\lambda)SK^+$ coefficients from the coefficients ranked in the first and the second half, respectively. Weights/connections corresponding to the sampled coefficients are reserved while others are deleted. We take $\lambda=0.75$ in all our experiments. In other words, connections from the half of higher correlations are three times as much as those from the half of lower correlations. The total kept connections/weights are $SK^+$, which depends on the degree of sparsity $S$.

The negative correlation coefficients are processed in a similar way, except that we consider the absolute value of the coefficients and keep more coefficients (and the corresponding connections/weights) of higher absolute values. The total sampled negative coefficients are $SK^-$, given $K^-$ negative coefficients from $r_{ik}$ for $k=1,2,\ldots,K$. Connections from each of output neurons $a_i$ are processed in the same way. Suppose there are $N$ output neurons $a_i$ for $i=1,2,\ldots,N$. Then the total sampled weights/connections are $SKN$. The dropping matrix $D$ are then created for training, in which $1$ denotes reserved weights and $0$ denotes deleted weights. $D$ has the same degree of sparsity $S$.

For convolutional layers, the set of correlation coefficients between neurons with shared connecting weights are jointly considered to determine whether a weight (or a set of connections with shared weights) should be reserved or deleted. Let $a_{im}$ be the $m$-th neuron in the $i$-th feature map of the current layer, and it is connected to $K$ neurons $b_{mk}$ in the previous layer for $k=1,2,\ldots,K$. ($K$ equals the filter size, \eg, $3\times3$, times the number of input channels.) The set of $K$ neurons $b_{mk}$ are determined by the position $m$. There are a total of $M$ neurons in the $i$-the output feature map as $a_{im}$ for $m=1,2,\ldots,M$. They all share the same set of $K$ weights, although connected to different sets of neurons in the previous layer $b_{mk}$ for $m=1,2,\ldots,M$. Weights between $a_{im}$ and $b_{mk}$ are shared for $m=1,2,\ldots,M$. We calculate the mean magnitude of the correlation coefficients between $a_{im}$ and $b_{mk}$ for $m=1,2,\ldots,M$ as

\begin{equation}
r_{ik} \triangleq \sum_{m=1}^M \left| \frac{E[a_{im} - \mu_{a_{im}}][b_{mk} - \mu_{b_{mk}}]}{\sigma_{a_{im}}\sigma_{b_{mk}}} \right| \ \textrm{.}
\end{equation}

Similar to the case in the fully- and locally-connected layers, given the degree of sparsity $S$, we select $SK$ mean correlation coefficients (and the corresponding weights) from the set of $K$ coefficients $r_{ik}$ for $k=1,2,\ldots,K$. We sort $r_{ik}$ in descending order and randomly choose $\lambda SK$ coefficients from the first half with higher values and $(1-\lambda)SK$ from the second half with lower values. Again we set $\lambda=0.75$ in all our experiments. The set of $K$ weights $r_{ik}$ for $k=1,2,\ldots,K$ are processed in the same way for all $i=1,2,\ldots,N$ (given $N$ feature maps in the current layer). The total sampled weights are $SKN$. The dropping matrix $D$ are created for training similar to that created in the fully- and locally-connected layers.

\section{Experiments}

Our weight sparsifying algorithm is evaluated on the LFW dataset \cite{huang2007a} and the YouTube Faces dataset \cite{wolf2011}, respectively, which has been extensively evaluated for face recognition in unconstrained conditions in recent years \cite{berg2012,chen2013,simonyan2013,cao2013,taigman2014a,sun2014a,sun2014c,sun2015,taigman2015,schroff2015}. \cite{lfw_result_page,youtubefaces_result_page} summarize all the results reported on these two datasets, respectively. All our models are trained on the same training set as has been used to train the previous state-of-the-art DeepID2+ models \cite{sun2015}. Therefore our algorithm can be directly compared with DeepID2+. The training set has approximately $290,000$ face images from $12,000$ identities. It also has a separate validation set of approximately $47,000$ face images from $2000$ identities for selecting the free parameters of algorithms such as  learning rates and for other training uses. In testing we evaluated both the tasks of face verification and face identification on the LFW dataset, as well as face verification on the YouTube Faces dataset. For face verification on LFW, $6000$ pairs of face images specified by LFW are evaluated to tell whether they are from the same person. For face identification on LFW, we follow the open- and closed-set protocols specified and evaluated by \cite{best-rowden2014,taigman2015,sun2015}. For face verification on YouTube Faces, $5000$ video pairs are compared, in which half are from the same person while the other half are form different people. After the face representations of our deep models are learned, Joint Bayesian algorithm \cite{chen2012,sun2014c} is used to learn a final metric for face recognition.

When training the baseline models, we use an initial learning rate of $0.01$, which is slowly decreased to approximately $0.0067$ after $140,000$ mini-batches of training. Then the model is fine-tuned for another $10,000$ mini-batches with ten-times smaller learning rates. There are $64$ pairs of faces in each mini-batch. 
After the baseline model $N_0$ is trained, we continue to train the sparsified models $N_1$, $N_2$, ..., $N_M$. As described in Tab. \ref{tab:algorithm}, each time only one additional layer are sparsified and the entire model are initialized by the previously learned model and re-trained. Since the model already has good initialization, we only re-train $70,000$ mini-batches when each time one additional layer is sparsified. The initial learning rate for re-training is the same as the that used to train the baseline model ($0.01$) but with a doubled decreasing rate. The last $10,000$ mini-batches among the $70,000$ mini-batches are also used for fine-tuning with ten times smaller learning rates. Joint identification-verification supervisory signals are used from the beginning of re-training.

\subsection{Sparsity improves performance}
\label{ssec:spa}

We test the face verification performance of our sparse ConvNets on the LFW dataset \cite{huang2007a} with various sparsity configurations in the fully-connected layer f, the locally-connected layers 5b and 5a, and the convolutional layer 4b (refer Tab \ref{tab:base} for the layer names). Our preliminary experiments show that the performance of a model is correlated with its total number of parameters from all layers. When the upper layers have already reduced a lot of the parameters, parameters in lower layers would become more critical and harder to reduce. The majority of parameters of our baseline model reside in the higher fully- and locally-connected layers. We reduce as many parameters as possible in these higher layers while the bottom convolutional layers are left untouched.

We use the degree of sparsity plus layer name to denote one sparsified layer of the model. For example, the fully-connected layer f with a degree of sparsity $1/256$ is denoted as $1/256$-f. When there are multiple sparsified layers, the uppermost layer is first sparsified. After re-training, the second highest layer is sparsified, and so forth. Tab. \ref{tab:sparse} shows a few configurations of our sparsified deep ConvNet models, the corresponding face verification performance on LFW, and the compression ratio (the number of parameters divided by that of the baseline model). Each column of the table is one particular configuration in which the sparsified layers and their degrees of sparsity are specified in the table. Layers not specified by the table are not sparsified and are leaved the same as those in the baseline model.

The second row of the table, in which the sparsity configurations are left blank, shows our baseline model with the face verification accuracy of $0.9895$ and the compression ratio is $1$. In rows 3-5, sparsity is gradually added from the topmost fully-connected layer f to the lower locally-connected layer 5b, and then to the convolutional layer 4b, with a consistent increasing of the face verification accuracy and decreasing of model parameters.

It is found that parameters in the fully-connected layer f and the locally-connected layer 5b are mostly redundant. The large number of parameters in these two layers actually hurt the model generalization ability. When parameters are dramatically reduced in these two layers (with extremely sparse $1/256$ and $1/128$ of the weights/connections of the baseline model, respectively), we improve the face verification performance of the original deep ConvNet from $0.9895$ to $0.9923$. When we further reduce half of parameters in the convolutional layer 4b, the accuracy further increases to $0.9930$, which improves $0.9895$ of the baseline model significantly, while the model parameters are reduced to $74\%$ of the original parameters.

Parameters in the locally-connected layer 5a are more critical than those in the higher layers 5b and f, although it has the most parameters in our baseline model. We find that parameters in layer 5a can only be made moderately sparse. When removing half of the connections in the locally-connected layer 5a (the second last row in Tab. \ref{tab:sparse}), we achieve $0.9922$ face verification accuracy with $43\%$ of the original parameters. To achieve a performance comparable to that of the baseline model, parameters in layer 5a could be reduced to $1/32$ of the original, while the total number of parameters is only $12\%$ of the original baseline model (the last row in Tab. \ref{tab:sparse}).

\begin{table}[t]
\small
\begin{center}
\begin{tabular}{p{4.0cm}|p{1.4cm}|p{1.4cm}}
\hline
sparse structure                & accuracy      & compression ratio  \\ \hline
                                & $0.9895$      & $1$                \\ \hline
$1/256$-f                       & $0.9898$      & $0.96$             \\ \hline
$1/256$-f $1/128$-5b            & $0.9923$      & $0.76$             \\ \hline
$1/256$-f $1/128$-5b $1/2$-4b   & $\bm{0.9930}$ & $0.74$             \\ \hline
$1/256$-f $1/128$-5b $1/2$-5a   & $0.9922$      & $0.43$             \\ \hline
$1/256$-f $1/128$-5b $1/32$-5a  & $0.9898$      & $0.12$             \\ \hline
\end{tabular}
\end{center}
\vspace{-4pt}
\caption{The LFW face verification accuracy and the number of parameters (normalized by that of the baseline model) for models with various sparsity configurations.}
\vspace{0pt}
\label{tab:sparse}
\end{table}


Since the locally-connected layer 5a has a dominating number of parameters in our model, we study how the performance degrades with respect to the total number of parameters in our model by changing the degree of sparsity in layer 5a while keeping other layers fixed. In particular, we take a sparsity configuration of $1/256$-f, $1/128$-5b, and $S$-5a ($1/256$ and $1/128$ degrees of sparsity in the fully-connected layer f and the locally-connected layer 5b, respectively, while changing the degree of sparsity $S$ in layer 5a). 
As shown in Tab. \ref{tab:degrade1}, the performance almost keeps with $26\%$ of the original parameters. It is still close to $99\%$ accuracy with only $12\%$ of the parameters.

\begin{table}[t]
\small
\begin{center}
\begin{tabular}{p{2.0cm}|p{1.7cm}|p{1.7cm}}\hline
$S$-5a  & compression ratio & accuracy  \\ \hline
$1$     & $0.76$            & $0.9923$  \\ \hline
$1/2$   & $0.43$            & $0.9922$  \\ \hline
$1/4$   & $0.26$            & $0.9918$  \\ \hline
$1/8$   & $0.18$            & $0.9908$  \\ \hline
$1/16$  & $0.14$            & $0.9890$  \\ \hline
$1/32$  & $0.12$            & $0.9898$  \\ \hline
\end{tabular}
\end{center}
\vspace{-4pt}
\caption{The LFW face verification accuracy when the sparsity of layer 5a (therefore the total number of parameters) changes in the model.}
\vspace{0pt}
\label{tab:degrade1}
\end{table}

\subsection{Correlation guided weight selection}
\label{ssec:cor}

We compare the correlation based weight reduction process described in Sec. \ref{ssec:drop} to random weight reduction, which is equivalent to setting the free parameter $\lambda$ (introduced in Sec. \ref{ssec:drop}, which balances the proportion of high and low correlation connections) to $0.5$. We have also investigated taking only connections (and the corresponding weights) with the highest correlations, by setting $\lambda$ to $1$, which is compared to the criteria adopted by our algorithm of selecting a majority of high correlation connections as well as a small portion of low correlation connections. We use additional letters r and h to denote random weight selection and the selection of the high correlation connections only, respectively. The experimented sparse structures and the comparison of the face verification accuracies on LFW are shown in Tab. \ref{tab:corr}. Our proposed connection selection criterion performs better than the other two criteria for various sparsity configurations.

We find that the mean absolute value of the correlations between neurons on the selected connections tend to increase after re-training on the sparsified structures. We calculate the mean neural correlations on connections selected either randomly or by our criterion before and after re-training. As shown in the last two columns of Tab. \ref{tab:corr}, the neural correlations increase in various degrees after re-training. This implies that highly correlated neurons are more helpful for prediction and re-training increase such correlations. Note that high correlations include both positive and negative correlations here.

\begin{table}[t]
\small
\begin{center}
\begin{tabular}{p{3.6cm}|p{1.1cm}|p{0.9cm}|p{0.9cm}} \hline
sparse structure                    & accuracy       & corr before & corr after \\ \hline
$1/256$-f                           & $\bm{0.9898}$  & 0.147 & 0.494 \\ \hline
$1/256$-f-r                         & $0.9893$       & 0.114 & 0.511 \\ \hline
$1/256$-f-h                         & $0.9893$       & & \\ \hline
$1/256$-f $1/128$-5b                & $\bm{0.9923}$  & 0.120 & 0.275 \\ \hline
$1/256$-f $1/128$-5b-r              & $0.9910$       & 0.089 & 0.272 \\ \hline
$1/256$-f $1/128$-5b-h              & $0.9902$       & & \\ \hline
$1/256$-f $1/128$-5b $1/2$-4b       & $\bm{0.9930}$  & 0.075 & 0.079 \\ \hline
$1/256$-f $1/128$-5b $1/2$-4b-r     & $0.9922$       & 0.067 & 0.073 \\ \hline
$1/256$-f $1/128$-5b $1/2$-4b-h     & $0.9925$       & & \\ \hline
\end{tabular}
\end{center}
\vspace{-4pt}
\caption{column 1-2: comparison of the LFW face verification accuracy for models with different connection/weight selection criteria, \ie, selection of a majority of high correlation connections (by default), random selection (denoted by letter r), and selection of the high correlation connections only (denoted by letter h). The comparison is conducted on the latest/lowest sparsified layers. Column 3-4: mean absolute value of neural correlations on selected connections on the latest/lowest sparsified layer before and after re-training.}
\vspace{-0pt}
\label{tab:corr}
\end{table}

We verify that a small portion of randomly selected connections besides those with the highest correlations helps to increase the complementarity of the neural predictions in the lower layer. Given a neuron in the current layer (the locally-connected layer 5b is used in this experiment), we find all neurons in the previous layer to which it connected to and calculate the correlations of neural activations between all pairs of neurons in the previous layer which are connected to the given neuron in the current layer. Tab. \ref{tab:random} reports the mean correlations of neural activations averaged over all pairs of neurons in the previous layer which are connected to a common neuron in the current layer. Neural connections in Tab. \ref{tab:random} are pruned by the criteria of 1) selecting connections with the highest correlations between the connected neurons (the second column); 2) selecting a majority of connections with the highest correlations and a small portion of randomly selected connections as we proposed (the third column); and 3) selecting connections randomly (the last column). Lower mean correlations in Tab. \ref{tab:random} indicate higher complementarity of neural predictions in the previous layer.

As can be seen in Tab. \ref{tab:random}, adding a small portion of randomly selected connections decreases the correlations between neural predictions in the previous layer, and therefore increases the prediction complementarity. However, further increasing the portion of randomly selected connections would hurt the performance due to the weakening of the predictive power of individual predictions.

\begin{table}[t]
\small
\begin{center}
\begin{tabular}{p{1.7cm}|p{1.5cm}|p{1.5cm}|p{1.5cm}}
\hline
            & highest corr  & high corr     & random    \\ \hline
mean corr   & $0.0914$      & $0.0860$      & $0.0820$  \\ \hline
\end{tabular}
\end{center}
\vspace{-4pt}
\caption{Mean correlation of neural predictions in the previous layer. See text for the detailed descriptions.}
\vspace{-0pt}
\label{tab:random}
\end{table}

\subsection{Why do we need a denser network?}
\label{ssec:why}

The results presented so far are surprising and also raise questions. Sparse networks with much fewer parameters outperform the dense one. The third row in Tab. \ref{tab:corr} shows that even after randomly removing most connections in the top fully-connected layer, the performance after re-training is comparable with the baseline model (0.9893 vs 0.9895). Then why do we need the denser baseline model? 

To answer these questions, Tab. \ref{tab:initial} reports the face verification accuracies of three sparse structures with random initializations being trained from scratch. It turns out that their performance is lower than the baseline model. The key difference is that our proposed algorithm adopts a layer-wise training scheme as described in Sec. \ref{sec:scnn}. Each time only one additional layer is sparsified and the entire model is re-trained. The initialization from weights learned in the baseline model is critical to continue learning the sparser models. Taking the same sparse structures, models learned from random initializations perform significantly worse than the properly initialized models, even much worse than our baseline model.

This result is interesting and has inspired our conjecture on the behavior of deep neural networks. Although the learning capacity of a sparse network is large enough to fit the training data, it is easier to get stuck at a local minimum, while a denser network with many more connections could help to find good initial solutions.
Once a good initialization is found by the denser network, sparsifying connections and re-training the network improve generalization. Tab. \ref{tab:initial} also compares the training-set face verification errors of the three models when they have finished training. Models initialized randomly have much larger training errors than those initialized by previously learned models. This implies that, without any prior knowledge, it is hard for a sparse network to find a good solution even on the training set. 



\begin{table}[t]
\small
\begin{center}
\begin{tabular}{p{3.3cm}|p{0.9cm}|p{1.1cm}|p{1.25cm}} \hline
sparse structure                & if pre-trained & accuracy & train error \\ \hline
$1/256$-f                       & yes & $0.9898$ & $0.0207$ \\ \cline{2-4}
                                & no  & $0.9887$ & $0.0229$ \\ \hline
$1/256$-f $1/128$-5b            & yes & $0.9923$ & $0.0302$ \\ \cline{2-4}
                                & no  & $0.9845$ & $0.0423$ \\ \hline
$1/256$-f $1/128$-5b $1/2$-4b   & yes & $0.9930$ & $0.0299$ \\ \cline{2-4}
                                & no  & $0.9833$ & $0.0463$ \\ \hline
\end{tabular}
\end{center}
\vspace{-4pt}
\caption{Comparison of face verification accuracies on LFW and face verification errors on our training set for sparsified models trained with or without  initialization from previously learned denser models.}
\vspace{-0pt}
\label{tab:initial}
\end{table}

\subsection{Method comparison}

We compare our neural correlation based weight pruning strategy with other pruning strategies proposed previously, including optimal brain damage (OBD) \cite{lecun1990}, weight magnitude based pruning (which is the bases of many weight pruning algorithms \cite{collins2014}), and Bayesian regularization and pruning (BRP) \cite{williams1994}. Comparison is conducted on the sparsification of neural connections in the fully-connected layer f, the locally-connected layer 5b, and the convolutional layer 4b, respectively, with the pre-specified degrees of sparsity of $1/256$, $1/128$, and $1/2$, respectively, as shown in the first column of Tab. \ref{tab:compare}. As did in Sec. \ref{ssec:spa} - \ref{ssec:why}, when pruning the layer 5b (or 4b), its previous layer f (or layers f and 5b) has already been pruned by our correlation based weight pruning algorithm.

For OBD, parameters with the largest saliency values defined by the second order derivatives of parameters are reserved, and then the sparsified model is re-trained. For the magnitude bases weight pruning, given the degree of sparsity $S$, $S\cdot|W^+|$ positive weights and $S\cdot|W^-|$ negative weights with the largest absolute values are reserved, in which $|W^+|$ and $|W^-|$ denotes the number of positive and negative weights, respectively. For BRP, weights are pruned iteratively under the L1 regularization, and each time weights with the smallest absolute values (less than $10\%$ of the mean absolute value in our implementation) are set to zero permanently. In the original implementation of BRP, learning rates are adaptable so that after each time of weight updating exactly one weight are vanished. However this is infeasible to large deep models as ours since there are too many weights to be pruned.

\begin{table}[t]
\small
\begin{center}
\begin{tabular}{p{1.25cm}|p{1.1cm}|p{1.25cm}|p{1.1cm}|p{1.25cm}} \hline
sparse structure    & OBD           & magnitude & BRP           & correlation   \\ \hline
$1/256$-f           & $\bm{0.9905}$ & $0.9888$  & $0.9863$      & $0.9898$      \\ \hline
$1/128$-5b          & $0.9903$      & $0.9912$  & $0.9902$    & $\bm{0.9923}$ \\ \hline
$1/2$-4b            & $0.9920$      & $0.9925$  & $0.9913$    & $\bm{0.9930}$ \\ \hline
\end{tabular}
\end{center}
\vspace{-4pt}
\caption{Comparison of different weight pruning strategies, including optimal brain damage (OBD) \cite{lecun1990}, weight magnitude based pruning (magnitude), Bayesian regularization and pruning (BRP) \cite{williams1994}, and our proposed neural correlation based pruning (correlation), on various sparse structures for face verification on LFW.}
\vspace{-0pt}
\label{tab:compare}
\end{table}

As shown in Tab. \ref{tab:compare}, our correlation based pruning strategy achieves the best performance for sparsifying layers 5b and 4b. Although OBD is better than our strategy in sparsifying layer f, its performance degrades significantly when pruning layers 5b and 4b, probably due to the difficulty of accurately estimating the second derivatives in lower layers.

\begin{figure}[t]
\begin{center}
\includegraphics[width = \linewidth]{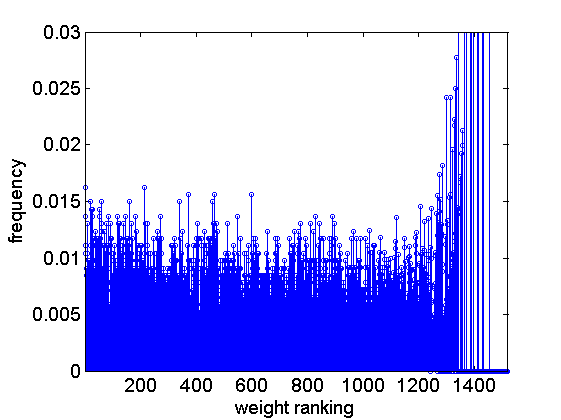}
\end{center}
\vspace{-10pt}
\caption{Distribution of weight magnitude rankings of weights selected by the proposed neural correlation based weight selection algorithm.}
\label{fig:weight_rank}
\end{figure}

We investigate whether correlations between connected neurons and weights on connections are correlated by counting the rankings of weights selected by our neural correlation based weight selection algorithm. For example, each neuron in the locally-connected layer 5b is connected to $3\times 3\times 256 = 2304$ neurons in the local regions of layer 5a. We rank the positive and negative weights on the $2304$ connections, respectively, by their magnitudes (absolute values), and keep the rankings of the reserved weights after correlation based weight pruning.  The frequency of the rankings of the reserved positive weights of all $3\times 2\times 256 = 1536$ neurons in layer 5b is counted in Fig. \ref{fig:weight_rank}. It can be seen that the rankings have a near uniform distribution, which means that large and small weights have equal chances of being reserved by our correlation based weight selection criterion. The rankings of the reserved negative weights have similar distributions. The same phenomenon is also found for layers 4b and f. The interesting phenomenon that neural networks with weights selected under a near uniform distribution of magnitude perform consistently better than weights selected with the largest magnitude indicates that weight magnitude is not a good indicator of the significance of neural connections.

\subsection{Sparse ConvNet ensemble}

We further verify our sparse ConvNet structures by training $25$ deep ConvNets taking a variety of face regions in different scales, positions, and color channels. We take the same $25$ face regions as used in \cite{sun2014c,sun2015}. The aim of training an ensemble of sparse ConvNets is to verify that the proposed sparse structure improves performance statistically, and also to construct a final high-performance face recognition system for evaluation.

We first train $25$ baseline models on the $25$ face regions. The model structures are the same as that shown in Tab. \ref{tab:base} except for the input dimensions. We take $112\times96$ input for rectangle face regions and $96\times96$ input for square regions. When input sizes change, feature map sizes in the following layers will change accordingly. After the baseline models are learned, we sequentially add $1/256$, $1/128$, and $1/2$ degrees of sparsity to the fully-connected layer f, the locally-connected layer 5b, and the convolutional layer $4b$, and refer the learned models as $1/256$-f, $1/128$-5b, and $1/2$-4b, respectively. The final sparsity configuration of the $25$ models is the same as that shown in the third last row in Tab. \ref{tab:sparse}. Adding sparsity to the specified three layers improves the mean face verification accuracy of the $25$ models by $0.18\%$, $0.17\%$, and $0.05\%$, respectively. The mean accuracy of the $25$ baseline models is $97.26\%$, and our proposed sparse structures improve the mean accuracy to $97.66\%$. Note that our baseline models already perform much better than DeepID2+. The latter has a mean accuracy of $96.61\%$ for $25$ models.

When combining features learned from the $25$ sparse ConvNets, we achieve $\bm{0.9955\pm 0.0010}$ face verification accuracy on LFW, which is better than $0.9947$ from the previous state-of-the-art DeepID2+ ensemble \cite{sun2015} given the same training data. FaceNet \cite{schroff2015} achieves an $0.9963\pm 0.0009$ face verification accuracy on LFW with approximated $700$ times the training data of ours. Given the face identification protocols adopted in \cite{best-rowden2014,taigman2015,sun2015}, we achieve $\bm{0.962}$ closed-set and $\bm{0.864}$ open-set face identification accuracies on LFW, respectively. Our result also improves the previous state-of-the-art $0.950$ and $0.807$ closed- and open-set face identification accuracies from DeepID2+ ensembles \cite{sun2015}.

In our further evaluation on YouTube Faces \cite{wolf2011}, our single sparse ConvNet achieves $92.7\%$ face verification accuracy on YouTube Faces, which is better than the $91.9\%$ face verification accuracy of a single DeepID2+ net \cite{sun2015}. The ensemble of $25$ sparse ConvNets achieves $93.5\%$ face verification accuracy on YouTube Faces, which is better than the $93.2\%$ face verification accuracy of the DeepID2+ ensemble \cite{sun2015}. YouTube Faces is a dataset harder than LFW due to the low quality video face images. There is a large domain gap between our training set, which contains high-quality static web images of celebrities, and YouTube Faces videos. Nevertheless, we achieve competitive performance on YouTube Faces, which verified the good generalization ability of our models.

\section{Conclusion}

This paper has proposed to learn effective sparser deep ConvNet structures iteratively from the previously learned denser models with a neural correlation based weight selection criterion. The denser model helps to avoid bad local minimums and provides good initializations which are essential for the sparser models to continue learning effective face representations, while the sparser model itself failed to learn effective features from data without the help of denser models. Empirical studies verified the superiority of neural correlations over weight magnitude or second order derivatives for selecting informative neural connections. The proposed sparse ConvNet with a moderate degree of sparsity ($26\%$-$76\%$ of weights in the dense model) significantly improved the performance of the original dense model, while the performance degrades slowly when the model further goes sparser. An ensemble of the proposed sparse ConvNet models achieved the state-of-the-art face verification/identification performance on the extensively evaluated LFW and YouTube Faces datasets.

{\small
\bibliographystyle{ieee}
\bibliography{egbib}
}

\end{document}